# INTELLIGENT INFORMATION EXTRACTION BASED ON ARTIFICIAL NEURAL NETWORK

Ahlam Ansari[1], Moonish Maknojia[2] and Altamash Shaikh[3]

Department of Computer Engineering, M.H. Saboo Siddik College of Engineering, Mumbai, India

*ABSTRACT*

*Question Answering System (QAS) is used for information retrieval and natural language processing (NLP) to reduce human effort. There are numerous QAS based on the user documents present today, but they all are limited to providing objective answers and process simple questions only. Complex questions cannot be answered by the existing QAS, as they require interpretation of the current and old data as well as the question asked by the user. The above limitations can be overcome by using deep cases and neural network. Hence we propose a modified QAS in which we create a deep artificial neural network with associative memory from text documents. The modified QAS processes the contents of the text document provided to it and find the answer to even complex questions in the documents.*

## 1. INTRODUCTION

NLP is crucial part of artificial intelligence. NLP can be used to understand the knowledge provided to the system and to process that knowledge [1]. NLP improvises the system and makes it behave human-like if programmed appropriately.

QAS is used for fast information extraction by exempting user to read unnecessary information which might not lead to the answer. For QAS there are various NLP models which are used. These models are symbol matching in which make use of linguistic annotation, structured world knowledge and semantic parsing[2] .There are various types of Symbolic matching models like frame-semantic parsing method, word distance benchmark method[3].These methods are not effective as they are only able to answer simple questions. The other method is using neural network.

Recent studies have shown that neural network's information processing mechanism is similar as human brain [4][5].Due to these similarities in functioning, it is used in various artificial intelligence models like pattern recognition, associative memory etc. and they yield high performance. There have been various methods to process natural language with neural network [6][7][8]. Success of these methods motivated us to implement QAS with help of neural network. In this paper we propose a method to create a deep neural network from the documents provide by the user and storing them for future use. We try to imitate human information recalling feature by processing the document first i.e. understanding it and then try to find answer to the questions asked. We process the question asked by the user and comprehending the question and understand what answer is required and then try to find the answer from the deep neural network created previously from the documents provided.

## 2. ARTIFICIAL NEURAL NETWORK

Artificial Neural Network (ANN) consists of interconnected processors called as neurons. Each neuron processes information and passes it to the next neurons. It was designed to operate similar to the human brain.

In this proposed QAS we use neural network for storing and retrieval of information. We try to imitate the information recalling feature of human by connecting the related information through links. The strength of the link between the two knowledge units is directly proportional to the relationship between them. The neural network represents the information understood by the system.

ANN has been used to solve various tasks that are hard to solve in the field of machine learning like speech recognition, handwriting recognition, and question-answering. ANN can be used to store the information previously processed and it can be used later for knowledge inference.

### 2.1. Natural Language Processing

NLP is a field of Artificial Intelligence, cognitive science and computational linguistics. It plays an important role for human-computer interaction. A computer understands only 1's and 0's and humans interact with each other by using natural language. For a computer to understand and process the language used by human NLP is required.

The document provided to the QAS contains data in natural language, to make QAS process the content of the document and create ANN; we will use NLP. We used the following NLP functions.

1. Part of Speech (POS) tagging
It is used to tag words from sentences with part of speech like noun, verb, adverb etc.
Example: "Popeye eats spinach"
Noun – Popeye and spinach.
Verb – eats.

2. Entity Recognition
It is used to label atomic elements in the sentence into categories like person, location, time, etc.
Example: "Gandhiji was born in Porbandar"
Person – Ghandhi
Location – Porbandar

### 2.2. Deep Neural Network

While processing natural language we can use the properties of Deep Neural Network (DNN). It contains various hidden layers. Such layers can help identify the relationship among various knowledge units while generating answer.

Two neurons might not appear to be related directly to each other in a simple ANN, but are connected indirectly through DNN. We can find such indirect relations if we construct a DNN for the same.

Suppose we want to find out if two people belong to the same village. We can assume that if there is blood relation among the two then they belong to the same village. The knowledge provided

International Journal in Foundations of Computer Science & Technology (IJFCST) Vol.6, No.1, January 2016

might not give sufficient information about the blood relation. ANN is constructed only by the data provided to the system. DNN can be used to go beyond the data provided to the system and identify the relationship among the two individual.

## 3. INTELLIGENT INFORMATION EXTRACTION SYSTEM

The proposed QAS consists of two phases namely learning and extraction. The system learns when new information is fed to it. While learning, new neurons (words) or links between neurons are added to the existing network. When a question is asked to the system it switches to the latter phase to extract the answer of the question asked. The block diagram of the proposed QAS is shown in Figure 1.

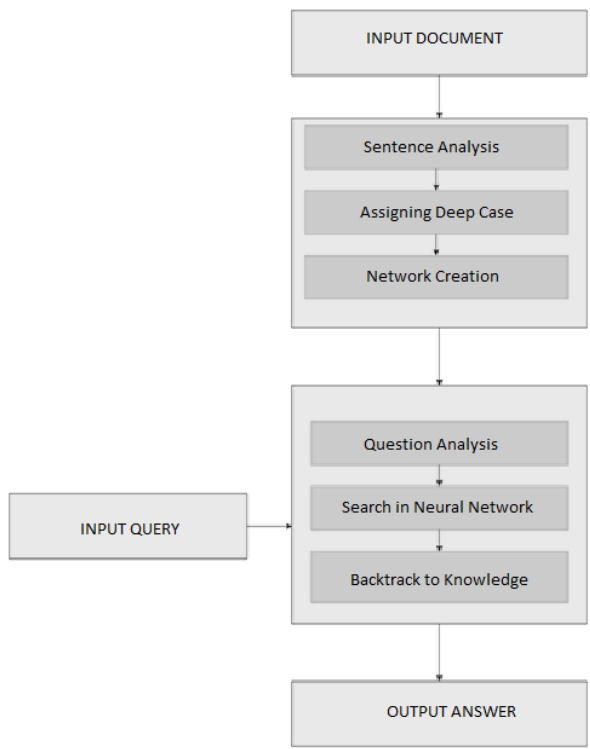

Figure 1: Block Diagram of QAS

### 3.1. Sentence Analysis

This step involves Natural language processing in which we derive knowledge units from sentence and words (neurons) from the knowledge unit.

### 3.2. Assigning Deep Case

We assign deep case to the neurons in artificial neural network to improve its inference capabilities. The deep cases that we are going to consider are shown below in Table 1.

Table 1. Deep Case

| Deep Case | Information |
|---|---|
| Agent | One who performs action or main actor |
| Action | The action(event) being performed |
| Location | Location where action occurred |
| Time | Time of event |
| Instrument | The object with which action is performed |
| Patient | Action being performed on. |
| State | Current condition/state of agent or patient. |

### 3.3. Network Creation

After sentence analysis we form an artificial neural network from the words extracted. If words exist in same knowledge unit we form a connection between them.

### 3.4. Question Analysis

In question analysis what answer is to be extracted can be estimated from the type of question asked, for e.g. if the question includes "where" then the answer will include a location. Words from the question are also extracted to search the neurons and the connection between them.

### 3.5. Search in Neural Network

The neurons (words) extracted from the question analysis is searched in ANN to find connection between them. This connection leads to knowledge layer.

### 3.6. Backtrack to Knowledge Layer

This step involves generating knowledge unit which possibly can be the answer of the question asked by user.

## 4. RESULT

We carried out experiment providing QAS a document which contains the details about Mahatma Gandhi.

Step 1: In this step we take input (text data) from the user. Let say document contains the following details.
"Gandhiji was born in Porbandar on 2nd October 1869".
Step 2: In this step we divide the input sentences into words and assign unique ID to each word. This division of the input data and ID assignment is shown in Table 2.

International Journal in Foundations of Computer Science & Technology (IJFCST) Vol.6, No.1, January 2016

Table 2: Assignment of id's to each word

| ID  | Words     |
|-----|-----------|
| Id1 | Gandhiji  |
| 1d2 | Born      |
| 1d3 | Porbandar |
| Id4 | 2-Oct-1869 |

Step 3: Now we extract knowledge units from the input data. Table 3 shows the extracted knowledge units and the ID's assigned to each unit.

Table 3: Assigning knowledge id's to each knowledge unit

| Knowledge ID | Knowledge |
|---|---|
| K1 | Gandhiji was born in Porbandar |
| K2 | Gandhiji was born on 2-Oct-1869 |

Step 4: In this step we assign word type to each word. The assignment of each word type is shown in Table 4.

Table 5: Assigning word type to each word.

| ID  | Words      | Word type |
|-----|------------|-----------|
| Id1 | Gandhiji   | Who       |
| 1d2 | Born       | What      |
| 1d3 | Porbandar  | Where     |
| Id4 | 2-Oct-1869 | When      |

Step 5: Now we assign deep cases to each word. Table 5 shows the various deep cases assigned to each word.

Table 6: Assignment of deep cases to each word

| ID  | Words      | Deep Case |
|-----|------------|-----------|
| Id1 | Gandhiji   | Agent     |
| 1d2 | Born       | Action    |
| 1d3 | Porbandar  | Place     |
| Id4 | 2-Oct-1869 | Date      |

Step 6: In this step we define the relationship between words and knowledge units. Row 1 in table 6 indicates that ID1 and ID2 are related because of knowledge unit K1.

Table 7: Relationship between knowledge units and words

| First Word | Second Word | Knowledge ID |
|---|---|---|
| ID1 | ID2 | K1 |
| ID2 | ID3 | K1 |
| ID1 | ID3 | K1 |
| ID1 | ID4 | K2 |
| ID1 | ID2 | K2 |
| ID2 | ID4 | K2 |

Step 7: Finally we create the network based on the sentence. The network created is shown in figure 2.

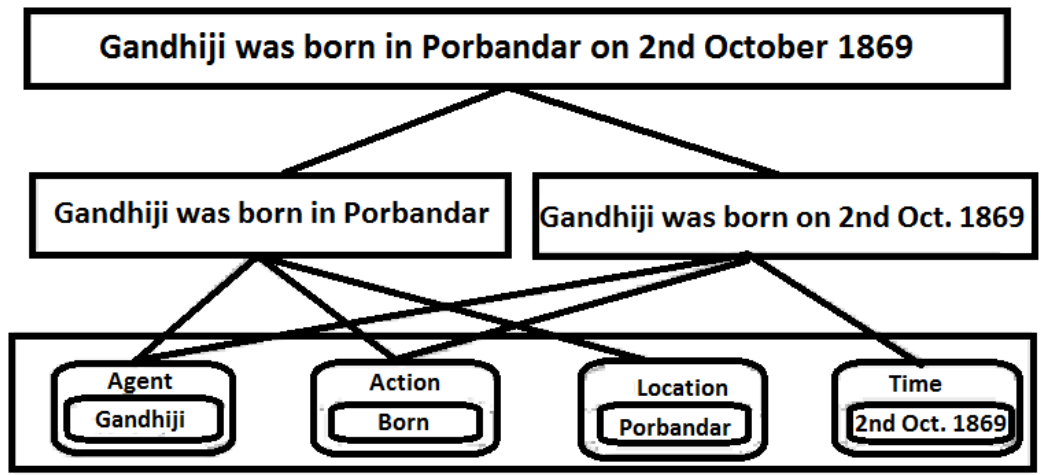

Figure 2: Network Diagram

Step 8: Now we have the neural network. The system has some knowledge and can answer questions based on this knowledge only. So we ask a question to the system.

"Where was Gandhiji born?"

We need to analyze each word from the question. Table 7 below contains the words in the question.

Table 8: Different words of the question

| Words |
|---|
| Gandhiji |
| Born |
| Where |

Step 9: Now we search the words from the network and extract the knowledge unit which relates these words. All the knowledge units that are related to the words in the question are shown in table 8.

Table 9: Searching the required knowledge units

| First Word | Second Word | Knowledge ID |
|---|---|---|
| **ID1** | **ID2** | K1 |
| **ID2** | ID3 | K1 |
| **ID1** | ID3 | K1 |
| **ID1** | ID4 | K2 |
| **ID1** | **ID2** | K2 |
| **ID2** | ID4 | K2 |

Here we have K1 and K2 which relates both the words '*Gandhiji*' and '*born*'.

K1 answers "where" whereas K2 answers "when". We evaluate this from the word type table that was created in learning phase. The user has asked "where" so the output knowledge unit will be K1.

Step 10: Now we have the required knowledge. So we output the selected knowledge unit i.e. K1 "*Gandhiji was born in Porbandar*"

## 5. CONCLUSION

The Proposed system overcomes the problems present in current systems to answer the complex question from the text document provided by the user. Proposed system solves this problem by assigning a deep case to each neuron which enhances the ability to answer more complex questions. Thus, user's time and efforts to find answer by reading the document will be minimized significantly.

### ACKNOWLEDGEMENTS

Our thanks to M.H. Saboo Siddik College of Engineering, Department of Computer Engineering, for giving us the initiative to do constructive work. We also thank anonymous reviewers for their constructive suggestions.